\documentclass[sigconf]{acmart}
\pdfoutput=1
\usepackage{booktabs} 
\usepackage{balance} 

\usepackage[
    type={CC},
    modifier={by-nc-nd},
    version={4.0},
    imagewidth={40pt},
]{doclicense}

\copyrightyear{2019} 
\acmYear{2019} 
\setcopyright{rightsretained} 
\acmConference[WSDM '19]{The Twelfth ACM International Conference on Web Search and Data Mining}{February 11--15, 2019}{Melbourne, VIC, Australia}
\acmBooktitle{The Twelfth ACM International Conference on Web Search and Data Mining (WSDM '19), February 11--15, 2019, Melbourne, VIC, Australia}
\acmDOI{10.1145/3289600.3291033}
\acmISBN{978-1-4503-5940-5/19/02}

\begin{document}
\title{Off-Policy Evaluation of \\ Probabilistic Identity Data in Lookalike Modeling}
\titlenote{All authors contributed equally to this work.}
\titlenote{Although the data used in this work is of a proprietary nature and will not be released publicly, Drawbridge Inc. is open to potential work with outside academic researchers. Those interested can reach out to randy@drawbridge.com}
\titlenote{\doclicenseLongText \newline \doclicenseImage}

\author{Randell Cotta, Mingyang Hu, Dan Jiang and Peizhou Liao}
\orcid{1234-5678-9012}
\affiliation{%
  \institution{Drawbridge Inc.}
  \streetaddress{2121 South El Camino Real, 7th Floor}
  \city{San Mateo} 
  \state{California} 
  \postcode{94403}
}
\email{{randy, mingyang, dan.jiang, peizhou.liao}@drawbridge.com}

\renewcommand{\shortauthors}{R. Cotta et al.}

\begin{abstract}

We evaluate the impact of probabilistically-constructed digital identity data collected between Sep. 2017 and Dec. 2017, approximately, in the context of Lookalike-targeted campaigns. The backbone of this study is a large set of probabilistically-constructed ``identities", represented as small bags of cookies and mobile ad identifiers with associated metadata, that are likely all owned by the same underlying user. The identity data allows to generate ``identity-based'', rather than ``identifier-based'', user models, giving a fuller picture of the interests of the users underlying the identifiers. 

We employ off-policy evaluation techniques to evaluate the potential of identity-powered lookalike models without incurring the risk of allowing untested models to direct large amounts of ad spend or the large cost of performing A/B tests. We add to historical work on off-policy evaluation by noting a significant type of ``finite-sample bias" that occurs for studies combining modestly-sized datasets and evaluation metrics based on ratios involving rare events (e.g., conversions). We illustrate this bias using a simulation study that later informs the handling of inverse propensity weights in our analyses on real data. 

We demonstrate significant lift in identity-powered lookalikes versus an identity-ignorant baseline: on average $\sim$70\% lift in conversion rate, CVR, with a concordant drop in cost-per-acquisition, CPA. This rises to factors of $\sim\!(4\!-\!32)$x for identifiers having little data themselves, but that can be inferred to belong to users with substantial data to aggregate across identifiers. This implies that identity-powered user modeling is especially important in the context of identifiers having very short lifespans (i.e., frequently churned cookies). Our work motivates and informs the use of probabilistically-constructed digital identities in the marketing context. It also deepens the canon of examples in which off-policy learning has been employed to evaluate the complex systems of the internet economy.

\end{abstract}

%
%



\maketitle
\newcommand{\Bias}{\textrm{Bias}}
\newcommand{\Var}{\textrm{Var}}
\newcommand{\CVR}{\mathrm{CVR}}
\section{Introduction}

The portion of the world economy devoted to marketing has grown much faster than the global economy itself. This is largely due to the rise of digital connectivity and the subsequent application of technologies that optimize the matching of brand messages with consumers. While powerful in principle, a proliferation of devices and maturation of web services has led to a complex and fractured view of consumers' interests, making it difficult to optimally match messages with consumers in practice.

Many of the largest companies in the world have become large precisely by solving this problem. They pair compelling free services with terms-of-use that allow the collection and monetization (via advertising) of vast amounts of rich personal data on large pools of users. Such use of personal data obviously has privacy implications and securing and guaranteeing the proper use of such data has proven challenging in practice \cite{kesan2015comprehensive,acquisti2016economics,landau2015control}. Given the tension between i) the need for companies to effectively reach customers and ii) the need to allow consumers options in terms of how their data is collected and used, a significant opportunity exists in solving this problem outside of this small set of massive platforms. 

The synthesis of \emph{Probabilistic Digital Identity Data} is one alternative solution to this problem. Here the idea is that sources of relatively ``thin" data coming mainly from the digital advertising ecosystem\footnote{This data can be seen as a massive stream of relatively uninformative event tuples (discussed later).} can be used to build inferred ``users" as collections of identifiers\footnote{By which we mean reset-able and often short-lived marketing identifiers like cookies and mobile ad-IDs. We refer to these simply as ``identifiers" in what follows.}. This organization can then be used to estimate the interests/behaviors of the ``users" who can, in turn, be messaged via their associated identifiers. As the name suggests, probabilistic digital identity data is created by leveraging statistical machine learning and data mining techniques, a route which is much more technically difficult compared to the deterministic aggregation of user-volunteered data around long lived identifiers (e.g., customerID). The data used to build the probabilistic identities in this work was passively collected and did not contain any information that would directly identify an individual such as name, phone number, or postal address. This kind of data is much less rich than the kinds of information typically volunteered to OSNs, thus, on the one hand, potentially lowering the risks associated with data breach/misuse, while on the other hand necessitating significant modeling to predict even basic demographic estimates (age, gender, etc.). 

While the usefulness of this data is conceptually intuitive to marketers, the impact of this data is often not straightforward to measure in practice. There are several different problem scopes for which one can consider evaluating the impact. There is also the issue of simply obtaining accurate estimates of effect size in the context of such complex systems. Likely owing to these factors, the authors know of no other work that has previously attempted to carry out these measurements. We attempt to do this here.

In terms of scope, we focus our evaluation of the impact of identity data on the important case of lookalike-targeted campaigns. Lookalike-targeting refers to the continual estimation of user interests and subsequent messaging of users that closely resemble those who have previously resonated with a given message. This is an interesting place to focus not only because of the strategy's general importance in marketing, but also because its analysis poses an intermediate level of difficulty\footnote{The importance of identity in re-targeting campaigns is obvious, but the analysis is not as interesting (amounting to a counting exercise), while an analysis for very broadly-targeted Brand campaigns is probably too difficult to do convincingly here due to the lack of signals to quantify the effects.}. To be concrete then, we are interested in answering the following question: \emph{If we allowed identity-powered user models to drive our lookalike targeting, instead of identity-ignorant variants, what would be the change in performance of the system?} By identity-ignorant lookalike-targeting we refer to systems based on user models that estimate the interests of the underlying user using only the data that is associated to a single identifier. Identity-powered systems, in contrast, model users' interests by aggregating data across all of the identifiers that are inferred to belong to the same underlying user.

To answer this question we lean heavily on off-policy evaluation \cite{precup2000eligibility}. The reasons for this are two-fold. First, we would like to avoid a) the risk of putting untested models into production systems that control large amounts of ad spend and b) the costs of performing many A/B tests to tune/validate new models. To the latter point, although A/B tests remain the gold standard for estimating the causal impact of interventions in principle, many have noted the often-impractical nature of carrying out these kinds of tests in practice (e.g., \cite{Joachims:2016:CEL:2911451.2914803,li2015toward,kohavi2012trustworthy}). Second, given that we will evaluate these models ``off-line", using only data that was previously collected by an ``on-line" model, it is important to correct a naive analysis to account for the fact that the naive results are generally not the same as would occur if the new model were actually allowed to make its own decisions and collect its own data. 

In the remainder of this paper, we will give a description of the raw data and the probabilistic processing that produces the identity data whose application we would like to assess the impact of. We will then discuss and analyze identity-powered lookalike targeting, using off-policy learning techniques to arrive at a reliable comparison of identity-powered and identity-ignorant systems. Finally, we discuss potential extensions of this work and conclude.

\section{Probabilistic Digital Identity Data}

The basic aim in generating probabilistic identity data is to organize massive sets of identifiers into small clusters such that the identifiers within a cluster are likely associated to the same human user. \textbf{Note in particular that the system cannot identify the devices or any other facts associated to one \emph{particular} individual, but rather, it simply attempts to determine that all identifiers in the cluster are likely owned by one person, and not two or more people}. From a technical perspective, the process involves aspects of information retrieval (what pairs of identifiers are even relevant to consider?), supervised learning (how likely is it that these two identifiers belong to the same person?), unsupervised learning (detecting users within a complex network of scored pairs of identifiers) and user-modeling (given a user cluster, what can we say about the user's interests?).

We first discuss the raw data upon which probabilistic identity data will be built, and then continue to discuss the processing logic that actually builds it. 
Further detail can be found in \cite{guan2016system}.

\subsection{Raw Data}

The raw data used to build identities comes from the platforms that power the digital advertising ecosystem. While this kind of data is massive in scale\footnote{There are more than $100$B of such events occurring daily across the globe.}, it comes as a stream of relatively uninformative, passively collected events. The signal-to-noise in such data falls well short of that typically volunteered by users to platform services like social networks, requiring significant processing to uncover even a basic knowledge of the underlying users and their interactions. 

The information inside of these events can be classified into two broad categories: i) observational data, typically including an identifier, timestamp and network address and ii) semantic data, describing the content context of the event (typically an associated content category\footnote{Like: \texttt{entertainment->movies}}, demographic estimates for the user, etc.). The identifiers associated to each event tuple are the cookies and/or mobile advertising identifiers used in the advertising industry to allow brands to reach consumers with their messaging. They are user-resettable unique identifiers (generated as large strings by cryptographic hashing algorithms). Active identifiers may thus become inactive at any time, new identifiers may spawn at any time, and many identifiers have a lifespan of only a single observation.

Given this kind of input data it is not even obvious which identifiers are associated to the same device, much less to the same user or to users living in the same household. We now discuss a probabilistic system that discovers this kind of identity structure from this unstructured raw data.

\subsection{Creating Identity Data}

The system that estimates the underlying identity structure from batches of raw data events consists of a chain of modules that are repeatedly applied in each run to build up hierarchical levels of organization (devices as collections of identifiers, users as collections of devices, households as collections of users). The basic components of the chain are: i) pair discovery, ii) pair scoring and iii) community detection.

\subsubsection{Pair Discovery}
The pair discovery step in the system is an information-retrieval step used simply to control the scale of what is processed by more expensive steps further downstream. Pair discovery performs a TF-IDF-like \cite{manning2008} scoring of identifier co-occurrences on bipartite identifier and ``proxy" networks, where the various proxy objects correspond to spatio-temporal localizations, for example, an \texttt{(IP address,date)} tuple. A conservative cut on this score allows the most relevant $\sim\!\!100$B pairs of identifiers to be considered further. 

\subsubsection{Pair Scoring}\label{subsubsec:pair-scoring}
In the pair scoring step supervised learning algorithms are trained to score the pairs produced by pair discovery according to obfuscated-deterministic labels. These labels are constructed using identifier-handle linkages in which the handles are obfuscated and typically refer to something like a hashed email. We label pairs in which the same handle is linked to both identifiers as $+1$'s, and those for which no handle is linked to both identifiers as $-1$'s\footnote{Given that both identifiers have \textit{some} handle/s attached. We label pairs in which one/both identifiers have no handles as $0$'s (this class composing the vast majority of pairs in the system).}. Given this, one can imagine that $+1$'s are pairs for which someone can verify that the same email has been used on both identifiers and $-1$'s are pairs for which we have failed to verify the same. We do expect that our $-1$'s are erroneous at some rate simply due to the imperfect knowledge of our label data providers and attempt to source identifier-handle linkages from multiple providers to reduce label errors that occur due to limited visibility.

Pair scoring, as a supervised learning task in exactly this context, has been the subject of prior work termed as ``cross-device'' linkage scoring. To date there have been multiple machine learning competitions \cite{icdm-drawbridge,cikm2016} regarding techniques for estimating which \texttt{id}s likely belong to the same underlying user \cite{diaz2015cross,landry2015multi,kim2015connecting,kejela2015cross,cao2015recovering,walthers2015learning,selsaas2015affm,lian2016cross,tran2016classification,phan2016cross}, among other works \cite{volkova2017cross,tanielian2018siamese,phan2017cross}. The features used in most pair scoring models, include those in the system described here, include several classes: i) those regarding the behaviors of each of the two objects in the pair in isolation ii) those having to do with the patterns of interaction between these two objects and iii) a featurization of the graph contexts in which this pair of objects is embedded\footnote{For graph embedding methods, see a recent review \cite{goyal2018graph} and the references therein.}. While features in ii) typically do the real heavy lifting, features in iii) are particularly important for accurately scoring pairs that do not already have a long history of interaction. While there are a handful of models involved in producing the features themselves, once produced we utilize an ensemble of boosted \cite{friedman2001greedy} and bagged \cite{breiman2001random} trees to produce the final scores.

\subsubsection{Community Detection}
If one views the set of scored pairs as a graph, this graph will be a web-scale complex network of weighted undirected edges. The graph has an approximately power-law degree distribution, small diameter and degree assortativity as in many other datasets \cite{newman2003structure}. In particular, simply removing edges with scores below some value will not reduce the graph into one in which the connected components are plausible representations of the true underlying users. For this we must apply a community detection (CD) algorithm \cite{fortunato2010community} to the graph of scored edges.

Given the scale of our graph, $\sim\!\!10$B vertices and $\sim\!\!100$B edges, the set of CD algorithms and implementations that can be effectively used is highly constrained. Perhaps the most scalable (and therefore most often used) massive CD algorithm is Label Propagation, LP \cite{raghavan2007near}. The extreme scalability of LP comes from its basis on a simple heuristic rule to create communities, however, and relying on such a rule has several downsides in our context: i) the basic rule leads to low-quality communities as measured both by eye and by truth data; ii) changing the basic rule is possible, but inconvenient as a mapping between tweaks to the rule and the resulting changes in performance is complex (one must reason about rules that may lend toward some particular objective, check if they, in fact, do improve this objective, etc.); iii) found communities are highly unstable to small input graph changes (clusters have little overlap in consecutive runs of the system). 

To overcome these issues the system that produces the identity data described here implements a highly-distributed local Fitness optimization procedure for its CD step that we term GFDC (Generalized-Fitness-Driven Clustering). The algorithm is related to many previous works \cite{lancichinetti2009detecting,lancichinetti2011finding,blondel2008fast,schaeffer2005stochastic,clauset2004finding}, though differing substantially in the definition of ``Fitness'' that is used. Historically, ``Fitness'' has tended to refer to the well-known network modularity \cite{newman2006modularity}, or to slight variants with free parameters \cite{lancichinetti2009detecting,lancichinetti2011finding} or by addition of other local network structural information \cite{de2015flexible}. For our purposes, with a network built on noisy and incompletely-observed links, pure modularity optimization does not always produce clusters that plausibly represent underlying users. For GFDC we suppose only that Fitness is a weighted sum of locally-calculable terms describing proposed communities. We use the Fitness function as a way to not only drive toward higher modularity, but also, for instance, to enforce desired behaviors like temporal smoothness \cite{chi2007evolutionary} or to provide Bayesian encouragement based on domain knowledge. As an example of the latter case, we can add a term that describes high-fitness user clusters as those in which there are about one (rarely zero or two) phones. Such a term would appear alongside a modularity term so that clusters having ten phones that might be proposed by modularity optimization alone would be broken down by the two-term definition of ``Fitness'' (provided the loss of modularity in doing so is not too great). 

The system uses Apache Giraph\footnote{http://giraph.apache.org/} to run GFDC at the required scale. The calculation proceeds in rounds, wherein each round a random selection of vertices can decide to join any adjacent community based on a calculation of the changes in Fitness that would result. The decision is greedy (vertices always choose the community that maximizes fitness) and myopic (vertices assume that no other vertices will move, which is not the case). Despite this, the flexibility of our definition of Fitness results in both greatly increased precision at all recall values with respect to an LP benchmark (measured by pairwise labels described in subsection \ref{subsubsec:pair-scoring}) and clusters that are much more reasonable-looking and stable. Of course, GFDC is also more expensive to run than LP, but given our distributed implementation, not prohibitively so, even at this scale.

\subsection{Probabilistic Identity Dataset}
Given that probabilistic techniques are being employed on noisy, incomplete, massive scale, dynamic data, we expect that any given set of putative ``identities" are afflicted with both Type-I errors (two identifiers placed in the same cluster which are not actually associated with the same user) and Type-II errors (two identifiers that are placed in two different clusters, even though they are actually associated to the same user). Any practical implementation of this technology allows tuning between the two ends of this error spectrum (small/tight clusters with very few False Positives (FP) and many False Negatives (FN) vs. large/fuzzy clusters with very few FNs and many FPs) to address use-cases with varying degrees of tolerance to the two types of error. In our system, after users are produced in the CD step, we give each identifier a cluster membership score based on simple measures of affinity calculated on the input network. 
Cutting on this score ``tunes'' our user clusters for a given FP/FN tolerance. In order to gauge the effect of this choice on the measurements reported below we present results for a handful of operating points along this spectrum of errors.

The identity datasets that we will use here have been created based on events generated during the period Sep. 2017 - Dec. 2017. After a basic cleaning to remove irregular-looking clusters the whole set is of size $\sim$900M identities, composed of a total of $\sim$23B advertising identifiers that were active at some time during this period. Basic statistics describing our high-FP and high-FN identity dataset endpoints are given in Table \ref{tab:graphdata}.
\begin{table}[!t]
\renewcommand{\arraystretch}{1.1}
\caption{Statistics for some of the Identity datasets used in this work. $G_1$ ($G_8$) is the operating point exhibiting the highest tolerance for FPs (FNs) that we consider here. Basic filters have been applied to remove clusters that are not plausible regular consumers (i.e., perhaps bots, etc.). Reported are figures for the number of users internationally and in the U.S. market specifically as well as the median number of identifiers per user and physical devices per user (US).}
\label{tab:graphdata}
\centering
\begin{tabular}{|c|c|c|c|c|}
\hline
Dataset & Global (M) & US (M) & ID/Usr (US) & Dev./Usr (US) \\\hline\hline
$G_1$ & $901$ & $264$ & $33$ & $3$ \\\hline
$G_8$ & $602$ & $176$ & $5$ & $1$ \\
\hline
\end{tabular}
\end{table}

\section{Identity-Powered Lookalikes}

In this section, we first discuss lookalike systems in more detail. We discuss the metrics that we will use to evaluate performance, data preparation procedures specific to this analysis and features/models employed to model users. We then turn to a detailed description of our off-policy evaluation methodology. We define basic concepts, lay out a simplified model of the larger system that allows to isolate the impact of identity and, finally, discuss a novel kind of bias that arises in the presence of rare-events and finite datasets.

\subsection{System, Models, Metrics and Data}

The job of a lookalike targeting system is to continually execute: i) collect historical impressions and conversions associated with a given campaign, ii) learn to discriminate between previously converted users and random users, iii) rank non-converted users according to their similarity with the previously converted group and iv) expose highly-ranked users to the campaign message by bidding appropriately to serve them impressions. As this loop is repeatedly iterated the pool of previously converted users grows and the system continually refines its ability to message users that particularly resonate with the campaign. 

\begin{table}[tb]
\renewcommand{\arraystretch}{1.1}
\caption{Basic statistics of the campaign data used here. Quoted for each campaign (Cmp) are the impression scale (Imp), conversion rate for IDs in and out of the production whitelist (CVR$_{\textrm{in}}$ and CVR$_{\textrm{out}}$, resp.), and descriptions of the campaign market vertical and call to action.}
\label{tab:campaigndata}
\centering
\begin{tabular}{|c|c|c|c|c|c|}
\hline
Cmp & Imp(M) & $\textrm{CVR}_{\textrm{in/out}}(\%)$ & Vertical & Action \\\hline\hline
A & 1.21 & 0.049/0.016 & Home Security   & Engagement\\\hline
B & 7.27 & 0.014/0.010 & Home Goods      & Form Fill\\\hline
C & 8.18 & 0.032/0.009 & Arts/Crafts     & Purchase\\\hline
D & 0.80 & 0.053/0.009 & Wireless        & Form Fill\\\hline
E & 1.46 & 0.018/0.009 & Streaming Video & Form Fill\\\hline
F & 0.45 & 0.027/0.008 & Solar           & Land'g Pg.\\\hline
G & 3.71 & 0.046/0.005 & Live Music      & Purchase\\\hline
H & 5.43 & 0.008/0.005 & Real Estate     & Form Fill\\\hline
\end{tabular}
\end{table}

In our system 10M highly-ranked identifiers are whitelisted for rather aggressive bidding, while all identifiers are also potentially impressed in less specifically-targeted channels. This is important in the context of off-policy evaluation: \textbf{all} identifiers have some chance to be impressed, so the support in covariate space is not completely collapsed onto that of the whitelisted group. We use gradient-boosted tree ensembles\cite{friedman2001greedy,ke2017lightgbm} to learn to discriminate previously-converted users from random users. The features we use are primarily semantic features based on the content associated with the user's events and the basic activity patterns of the identifiers composing the ``user''.

Albeit basic in terms of machine learning algorithm and features, the aforementioned system serves as a qualified testing ground for the purpose of evaluating our identity data: The only difference between the identity-powered and identity-ignorant models is whether or not the same set of features is aggregated across all of the identifiers inferred to be associated to the same underlying user. Moreover, due to the generic nature of our system, we expect that the effectiveness of identity data found here would generalize well to other similar but more sophisticated targeting systems.

We focus on CVR as our measure of the performance difference between the two models as we observe similar results (both qualitatively and quantitatively) for CPA as we do for CVR. We filter the data such that only the first conversions and impressions occurring before the first conversion are counted for a given brand/user history, avoiding potential confusion caused by multiple conversions.

We chose a set of eight campaigns to include in the study, which could be useful in understanding the heterogeneity of our results. While the brands involved will remain anonymous, we note that the campaigns were chosen based on several criteria: i) impression and/or conversion scale large enough to reasonably study, though ranging from small to large within that criterion, ii) broad targeting (geographical region, demographics, etc.), and iii) collectively representing a variety of products and calls to action. Basic statistics describing the campaign data are provided in Table \ref{tab:campaigndata}.

In all analyses below the identity data used was created based on events that were generated before these campaigns were run. This eliminates the possibility of any leakage of the campaign results via the structure of the identity data itself.

\subsection{Off-Policy Evaluation}

Naively, to gauge the impact of identity on user-modeling, one simply trains two lookalike models (identity-powered and identity-ignorant versions) in an offline fashion, makes predictions about which users will convert, waits for the production (on-line) system to collect new data and compares their ability to predict which users did eventually convert. This does not answer our real question of interest: if we were to use either of these models \emph{in place of the production system} (part of which includes its own on-line lookalike model that does not use identity data), how would the results differ? New models would evaluate users, adjust bids and serve impressions differently than the online model. To overcome this bias one should re-weight the data collected by the production system to bring estimates into closer accord with what would be expected from alternative systems.

\subsubsection{Inverse-Propensity Weighting}
A procedure which is commonly used to do this correction is \emph{importance weighting} \cite{horvitz1952generalization} or \emph{inverse-propensity weighting} \cite{hirano2003efficient} (``IPW", in the case where propensity models are being used to work in high-dimensional covariate spaces). In equations this can be seen as a change of measure on covariate space:
\begin{equation}\label{eqn:estimate_Y_star}
Y = \int_{\vec{x}} y P(\vec{x}) \quad\! \textrm{and}\quad\! Y^* = \int_{\vec{x}} y P^*(\vec{x}) = \int_{\vec{x}} y \left(\frac{P^*(\vec{x})}{P(\vec{x})}\right)P(\vec{x}),
\end{equation}
where $Y$ and $Y^*$ are the outcome quantity of interest (CVR) estimated under the on-line or off-line policy, respectively ($Y^*$ being the main subject of our interest here). $\vec{x}$ is an example point in covariate space. $P(\vec{x})$ and $P^*(\vec{x})$ are the sampling distributions representing systems leveraging the on-line and off-line models, respectively. For our purposes, $P(\vec{x})$ will be defined as the probability that an ad request represented by covariates $\vec{x}$ is impressed under the on-line bidding policy. The estimates for $Y$ and $Y^*$ is then
\begin{equation}\label{eqn:estimate_Y_star_numeric}
\hat{Y} = \frac{1}{n}\sum_{i\sim P} y_i \quad\! \textrm{and}\quad\! \hat{Y}^* = \frac{1}{n}\sum_{i\sim P} \left(\frac{P^*(\vec{x}_i)}{P(\vec{x}_i)}\right) y_i = \frac{1}{n}\sum_{i\sim P} w_i y_i,
\end{equation}
where $\sum_{ i\sim P}$ means summing over data points $i$ that are sampled from distribution $P$.
Then, we see that estimating the quantity of interest $\hat{Y}^*$ requires up- or down-weighting examples collected under the on-line policy according to weights $w=P^*(\vec{x})/P(\vec{x})$. Examples that would be more commonly seen if the off-line model had been used to collect data rather than the on-line model that was actually used, i.e., when $P^*(\vec{x})>P(\vec{x})$, are thus up-weighted (and vice-versa).

\setlength\belowcaptionskip{-6pt}
\begin{figure}[tb]
\includegraphics[width=0.40\textwidth]{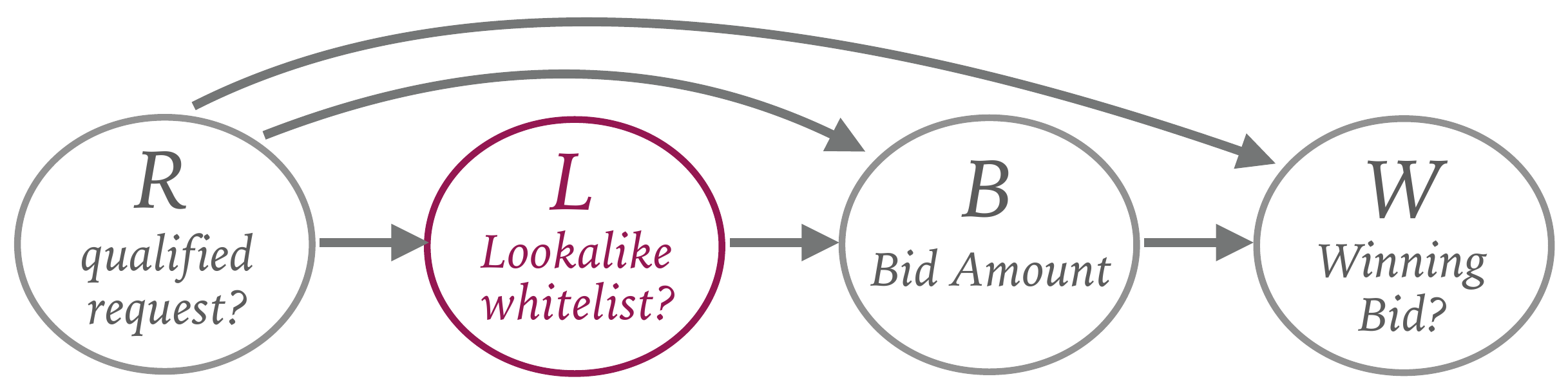}
\caption{Approximate factorization of our system bidding logic as concerns the Lookalike Model. We study a direct change to the variable $L$, which indirectly changes the variable $B$. Both $R$ and $W$ remain approximately unchanged.}
\label{fig:StructuralModel}
\end{figure}
Given that we are changing only one piece in a larger complex system, one can usefully employ a structural model \cite{wright1921correlation,pearl1998graphs} to factorize the larger $P(\vec{x})$ and $P^*(\vec{x})$ distributions such that many of the factors cancel in their ratio. In our bidder, the influence of the Lookalike model can be isolated using the following Markov factorization (corresponding structural model pictured in Fig. ~\ref{fig:StructuralModel}):
\begin{eqnarray}
P(\vec{x})&=&P(R,L,B,W) \\
&\approx& P(W|B,R) \cdot P(B|L,R) \cdot P(L|R) \cdot P(R)
\end{eqnarray}
Both of the factors $P(R)$ and $P(W|B,R)$ will cancel out in the ratio, corresponding to the statements that i) The characteristics of requests that pass basic targeting criteria for some campaign (``qualified requests") is independent of the internal model logic that we will apply in our system and ii) conditioned on the the request $R$ and the bid amount $B$, the external auction is isolated from knowledge of what model we applied to come up with that bid amount. Both of these statements are approximate, but expected to hold quite well. The former is true up to changes in the nature of internal competition between the various campaigns in our own system, and the latter is true up to changes in the auction environment as external competitors adjust to changes in our bidding behavior.

In contrast to $P(R)$ and $P(W|B,R)$, we do not expect the factors $P(B|L,R)$ and $P(L|R)$ to cancel out in the ratio. The factor $P(L|R)$ is precisely what we are changing when we change the Lookalike model, it is the probability of a given user being placed on the Lookalike whitelist with respect to some campaign. Getting on this list depends primarily on the model score (the part we will change by using identity-powered features rather than identity-ignorant features), but also secondarily on seemingly un-involved covariates regarding user activity level and behavior patterns\footnote{We do not consider all possible users for Lookalike whitelisting, the considered set partly depends on how active the users are, etc.}. As there are a handful of such covariates, we will employ simple logistic regression models to estimate $P(L|S,X)$, $P^*(L|S^*,X)$, where $X\subset R$ are activity/behavioral covariates and $S$ and $S^*$ are identity-ignorant and identity-powered Lookalike scores, respectively. 

The bid-amount term, $P(B|L,R)$, is also interesting to consider. In our system bid amounts are largely influenced by feedback of outcome values (i.e., bids tomorrow are a function of CVR and CPA today, among other considerations), so, strictly speaking, there is a feedback loop that should be included in our structural model which would invalidate its use in causal studies \cite{pearl2009causality}. Given this, we perform two analyses, where we set the ratio of these factors as:
\begin{eqnarray}
\textrm{i)}\quad\frac{P^*(B|L,R)}{P(B|L,R)}=1 \quad\textrm{and,}\quad\textrm{ii)}\quad \frac{P^*(B|L,R)}{P(B|L,R)}=\frac{P^*(L|S^*,X)}{P(L|S,X)}.
\label{bidfactors}
\end{eqnarray}
For the first case we set the ratio $=1$ (canceling the bid term), which will give estimates for the performance that we should expect in the time before which feedback is incorporated\footnote{In this case, this is $\sim$days, as conversion feedback is sparse and latent data.}, or in the case where feedback is not the primary driver of bid prices. In the second case, we make the following series of assumptions: that the ratio of the bid amounts would scale with the ratio of CVRs for the two models, which itself would scale with the ratio of the two propensity scores. Of course, both of these choices are rough approximations, but useful inasmuch as we expect the first to under-estimate and the second to over-estimate the true steady-state IPWs. We thus expect the true performance of a deployed Lookalike model to lie somewhere between the two estimates obtained in what follows.

\begin{figure}[tb]
\includegraphics[width=1.\columnwidth]{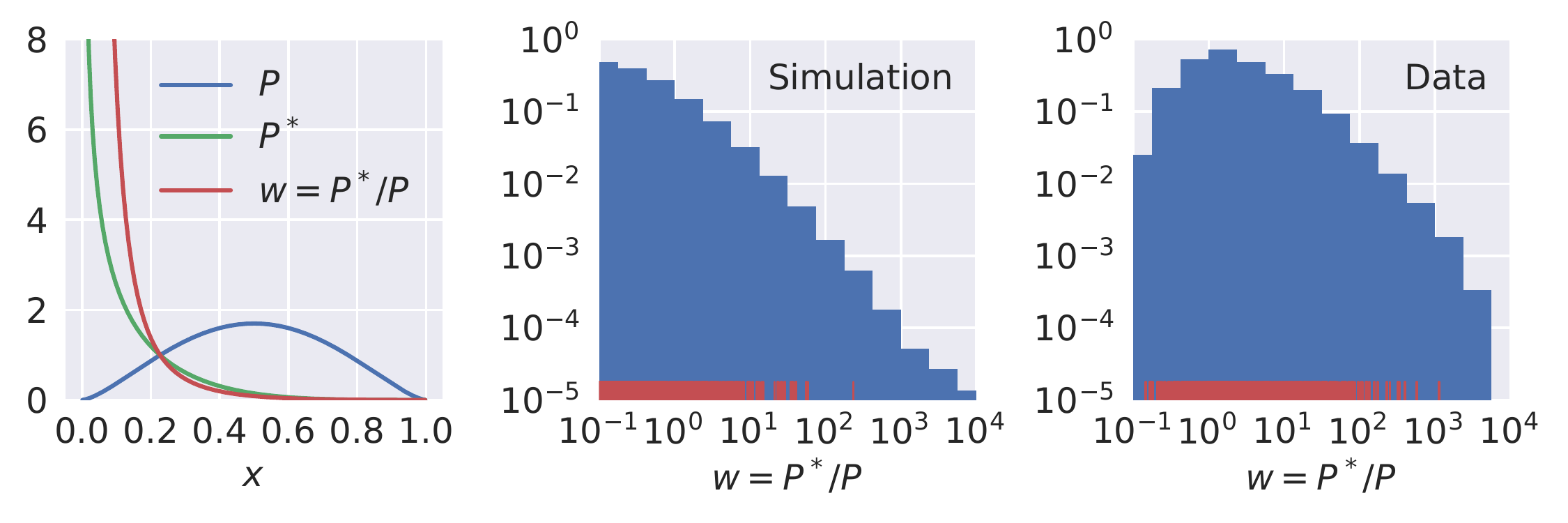}
\caption{The sampling distributions $P$ and $P^*$ in the simulation study are shown on the left. The histogram of the resulting IPW is displayed in the center panel, which qualitatively matches the histogram from real data shown on the right. Conversions are plotted as red bars on the x-axis in both histograms, showing the regions in which they are rare.}
\label{fig:toy-model}
\end{figure}
An important consideration in off-policy evaluation is in how to treat examples falling in regions of support non-overlap, i.e., examples near which the IPW, $w$, explodes. In such regions a naive weighting by the IPW implies that a few examples dominate the estimation, greatly increasing the variance of our estimates. Typical heuristic solutions trade off most of the variance at the cost of some bias. 
Here we use weight truncation \cite{li2015offline}, $w_T(w)=\min(w,w_0)$, 
where weights are truncated as they cross a threshold $w_0$. While many other works \cite{jiang2016doubly,gilotte2018offline} have focused on comparing a wide variety of counterfactual estimators (e.g., weight dropping, truncation, model-based and doubly-robust variants) we opted not to do this here, as our main focus is on the evaluation of the impact of identity data, and as we were not able to follow up with the kind of extensive A/B testing that would be required to definitively judge the relative performance of various estimators. We did do our analysis twice, once using the weight truncation mentioned above, and once using weight dropping \cite{2012arXiv1209.2355B}, where weights are set to zero upon crossing a threshold $w_0$, finding estimates that were consistent with $w_T$ and of similar variance (these are omitted for clarity in what follows).

\subsubsection{Finite-Sample Bias}
Besides the bias/variance trade-off that usually determines $w_0$, in our analysis we encountered another kind of bias. The issue here is caused by a confluence of factors: the sparse nature of conversions, the modestly-sized (in terms of conversions) campaign datasets used here, and the importance of the upper tail of the IPW distribution. Given the sharply falling tail of the IPW distribution, for some region $w_0>w_s$ one statistically expects only a small amount of impressions and thus very few conversions, even though the CVR in that regime may well be quite high. This regime of $w_0>w_s$ is pertinent because it is precisely where we hope the new model to improve upon the old, yet we could be counting thousands of highly-weighted impressions and the estimated CVR collapses to unreasonably low values. We refer to this source of bias as \emph{finite-sample bias}.

A simple simulation study illustrates this effect. It is also a nice sandbox in which practitioners can learn basic rules for setting $w_0$ in analyses on real data. Here we suppose a population of identifiers indexed simply by $x \in [0, 1]$. Each member of the population has an inherent CVR, $y(x)$, which is linear in $x$ for simplicity. We suppose that the production and new models have true propensity functions of $P(x)$ and $P^*(x)$, respectively, that are Beta distributions (illustrated in Fig.~\ref{fig:toy-model}), with parameters chosen to match the qualitative behaviors of our actual data, especially in the high-IPW tail\footnote{The actual distributions were: $P(x) \sim B(2.5, 2.5)$ and $P^*(x) \sim B(0.5, 4.5)$.}.

\begin{figure}[tb]
\includegraphics[width=.99\columnwidth]{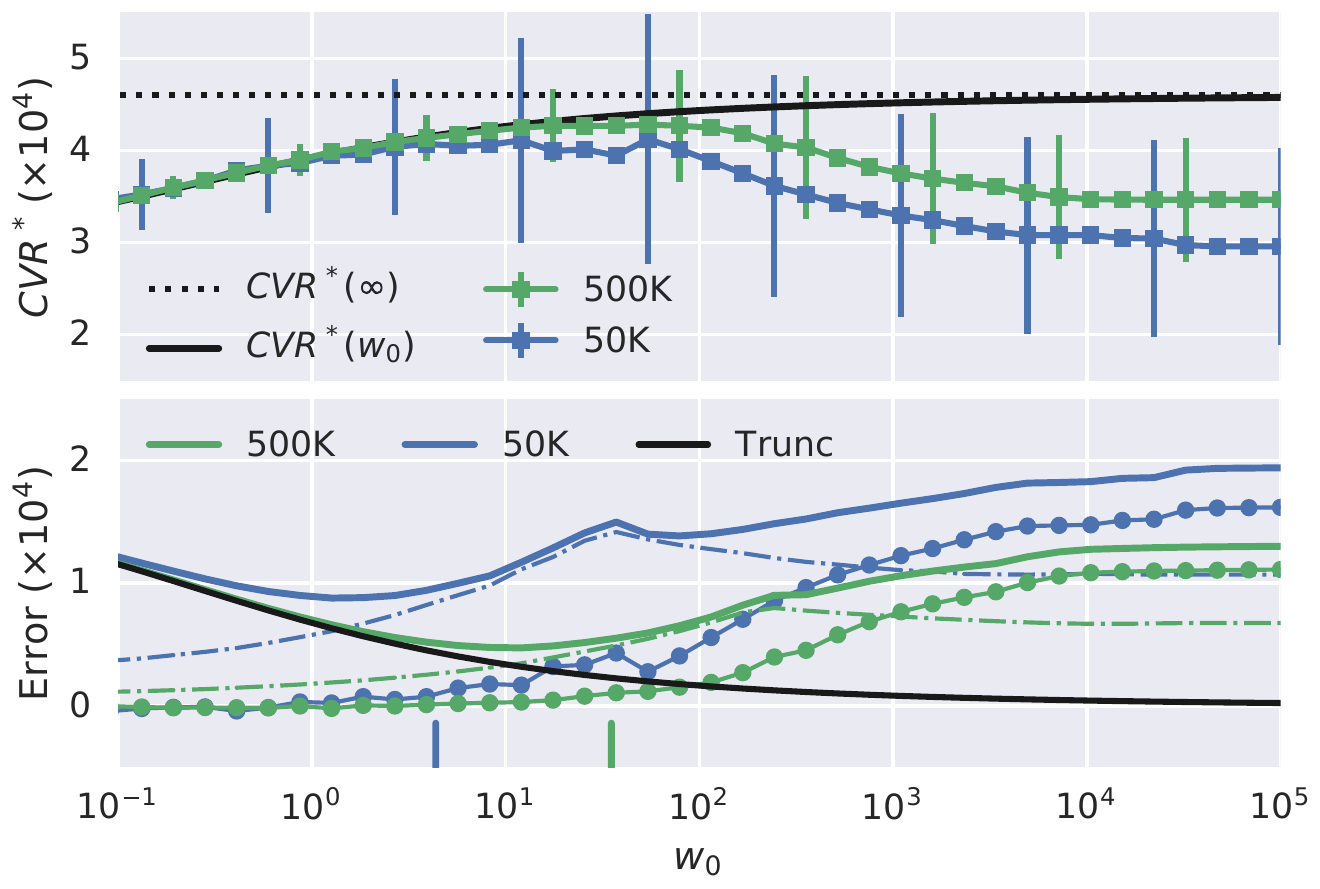}
\caption{Top, $\mathrm{CVR}^*$ as a function of IPW threshold $w_0$, is calculated (squares) from two batches of datasets of 50K (green) and 500K samples (blue), and compared to infinite-sample $\mathrm{CVR}^*(w_0)$ (solid black) and ground truth $\mathrm{CVR}^*(\infty)$ (i.e. no truncation, black dotted). Bottom, the total error (solid color) is decomposed into the finite-sample bias (spheres), truncation bias (solid black) and standard deviation (dash-dot). The heuristic choice of $w_0$, Eqn.~\ref{eqn:w_0_condition}, is shown as vertical bars on the x-axis.}
\label{fig:toy-model-results}
\end{figure}

\begin{figure*}[th]
\includegraphics[width=0.99\textwidth]{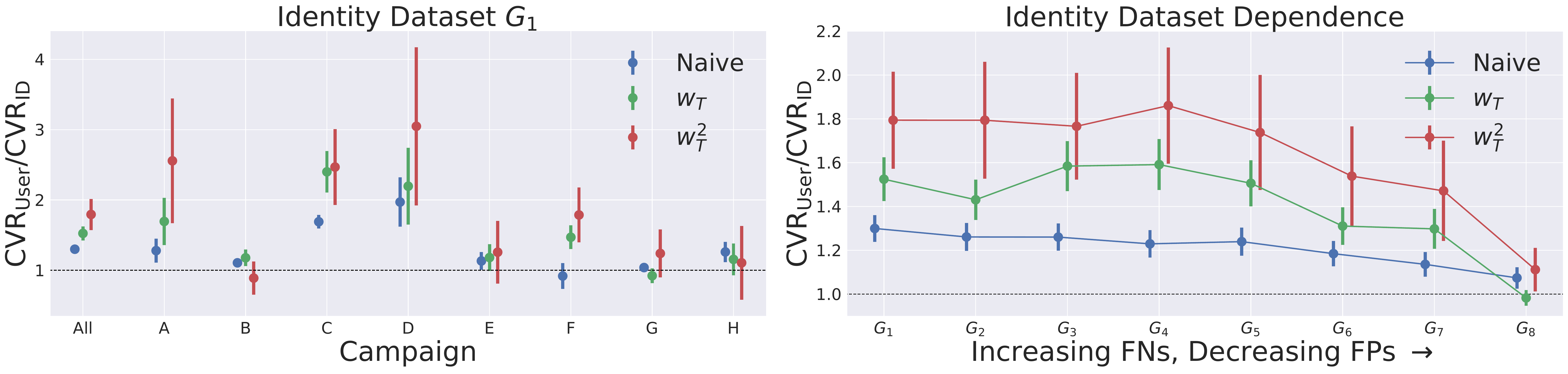}
\caption{Estimates for the lift in CVR for lookalike-targeted campaigns due to the use of identity data. In the left panel, results by-campaign and averaged over campaigns for the $G_1$ (high-FP) identity dataset. On the right, campaign-averaged results for a series of identity datasets ranging between the $G_1$ and $G_8$ endpoints. Naive estimates are presented along with corrected results using overall weights of $w_T$ or $w_T^2$, according to the choice of bid-factor (c.f. Eqn. \ref{bidfactors}).}
\label{fig:UserModelResult1}
\end{figure*}

Given this setup we can estimate $\CVR^*$ both analytically (c.f. Eqn.~\ref{eqn:estimate_Y_star}) and numerically (c.f. Eqn.~\ref{eqn:estimate_Y_star_numeric}), allowing a knowledge of infinite-sample and finite-sample results, respectively, and thus a decomposition of the overall bias, ${\rm Bias}_{\rm all}$, into truncation, ${\rm Bias}_{\rm T}$, and finite-sample, ${\rm Bias}_{\rm FS}$, components as following:
\begin{eqnarray}\label{eqn:biases-breakdown}
{\rm Bias}_{\rm all} &=& \widehat{\CVR}^*(w_0) - \CVR^*(\infty) \nonumber \\
{\rm Bias}_{\rm T} &=& \CVR^*(\infty) - \CVR^*(w_0) \\
{\rm Bias}_{\rm FS} &=& {\rm Bias}_{\rm all} - {\rm Bias}_{\rm T} \nonumber
\end{eqnarray}
In other words, the overall bias is the difference between the estimated $\widehat{\CVR}^*(w_0)$ and the ground truth $\CVR^*(\infty)$; the infinite-sample truncation bias equals the ground truth minus the true $\CVR^*$ with truncation, \emph{analytically} calculated by integrating Eqn.~\ref{eqn:estimate_Y_star} over the range $[0, w_0(x)]$; the finite-sample bias is the leftover bias after accounting for the truncation.

The one term left in Eqn.~\ref{eqn:biases-breakdown}, $\widehat{\CVR}^*(w_0)$, needs to be numerically calculated. Here, 1) one dataset $i$ with $50$K or $500$K impressed identifiers is sampled from the known Beta distribution $P(x)$. 2) It is then boostrapped 100 times, allowing us to calculate the median and variance of $\widehat{\CVR}^*$ at different $w_0$ \emph{for this dataset $i$} in the same way a normal practitioner would do. Note that, without the knowledge of the underlying probability distribution $P$, the estimates $\widehat{\CVR}^*(w_0)_i$ when $w_0\gg 1$ would not be robust due to the relatively large variance in the modest-sized data, and the variances $\Var(w_0)_i$ would be underestimated from bootstrapping\cite{dasgupta2008asymptotic}. In order to obtain a robust and accurate estimate of $\widehat{\CVR}^*(w_0)$ for the bias breakdown (Eqn.~\ref{eqn:biases-breakdown}), and to correctly quantify the variance so that the total error $({\rm Bias}_{\rm all}^2 + \Var)^{\frac12}$ can be measured, step 1) and 2) are repeated 100 times, and the median of these ${\widehat{\CVR}^*(w_0)_i}$ and ${\Var(w_0)_i}$ are used as $\widehat{\CVR}^*(w_0)$ and its variance. 

The results from the simulation study are shown in Fig.~\ref{fig:toy-model-results}. 
We observe the expected trade-off between truncation bias at low $w_0$ and variance at higher $w_0$. At even higher $w_0$ there is the regime with large finite-sample bias (larger for smaller datasets, as expected). Practitioners with real datasets may expect to see an estimator whose variance rises continually with $w_0$ and whose mean plateaus, but this is not the case: one observes the variance rising until there are no more conversions with higher IPW, around $w_0\approx 100$, at which point $\widehat{\CVR}^*$ collapses (with compulsory decrease in variance) and eventually falls significantly below the ground truth. Obviously, the optimal choice for $w_0$ is to minimize the total error shown in the lower panel in Fig.~\ref{fig:toy-model-results}. But this is not visible to a practitioner with a single real dataset and no knowledge of the underlying model. We, therefore, look for a simple heuristic that yields approximately optimal results in our toy model.

This choice should obviously depend on both the campaign size and nominal CVR. We find the following heuristic to provide good results in simulation studies:
\begin{equation}\label{eqn:w_0_condition}
    Pr(k_{>w_0}\leq 5; N_{>w_0}, \bar{Y}) = 1 - \delta,
\end{equation}
where $k_{>w_0}$ is the number of conversions and $\mathrm{N}_{>w_0}$ the number of impressions falling in the IPW region $w>w_0$, $\bar{Y}$ is the nominal CVR, computed as the average in the production system. The number of conversions thus follows a binomial distribution with parameters $N_{>w_0}$ and $\bar{Y}$ and we set $\delta=0.05$ so that we start truncating weights where we expect $\leq 5$ conversions at IPWs this extreme with $95\%$ probability. This choice is illustrated in Fig.~\ref{fig:toy-model-results} with vertical bars on the x-axis, which does a reasonable job at minimizing the total error for datasets of both sizes. We use this heuristic to choose $w_0$ in what follows.

\subsection{Results}

The basic results of our lookalike analysis are presented in Fig.~\ref{fig:UserModelResult1}, where we present campaign-specific and campaign-averaged results for the high-FP endpoint identity dataset $G_1$ as well as campaign-averaged results for identity datasets interpolating between our $G_1$ (high-FP) and $G_8$ (high-FN) endpoints.

Though the variance of our estimates is not negligible, we observe (Fig.~\ref{fig:UserModelResult1}, left) that the identity-powered lookalike system appears to show a significant $\sim\!70\%$ lift on average\footnote{This is a number bracketed by the $w_T$- and $w_T^2$-style off-policy corrected estimates (c.f. Eqn.~\ref{bidfactors}), averaged across campaigns. The Naive estimate of the same had lift $\sim\!30\%$.}. There are some campaigns for which the naive and corrected results largely agree. We find that these are generally those in which the identity-powered and identity-ignorant lookalike whitelists have significant overlap. In most cases where naive and corrected results significantly disagree the corrected results are larger, evidently showing that regions in covariate space that are preferred by the identity-powered model, are preferentially those with higher CVR.

Results comparing different identity datasets are also interesting (Fig.~\ref{fig:UserModelResult1}, right). Here we have some logical expectations for the shape of the lift curve: at very high FNs we should have very small clusters and the lift relative to identity-ignorant approaches should approach 1 (no lift); moving in the opposite direction we expect lift to rise as useful information is included, and then degrade once clusters become so large that the information being included is mostly noise. We indeed observe an increase in relative lift from nearly 1 (no lift) at the $G_8$ endpoint to a plateau with a higher lift as we approach the $G_1$ endpoint here, suggesting that our $G_1$ dataset, although admitting the most FPs of any here, has not reached the point of significant degradation. Unfortunately, our system does not produce identity datasets with higher FP rates than $G_1$, which we plan to investigate in the future. 

\begin{figure}[tb]
\includegraphics[width=0.99\columnwidth]{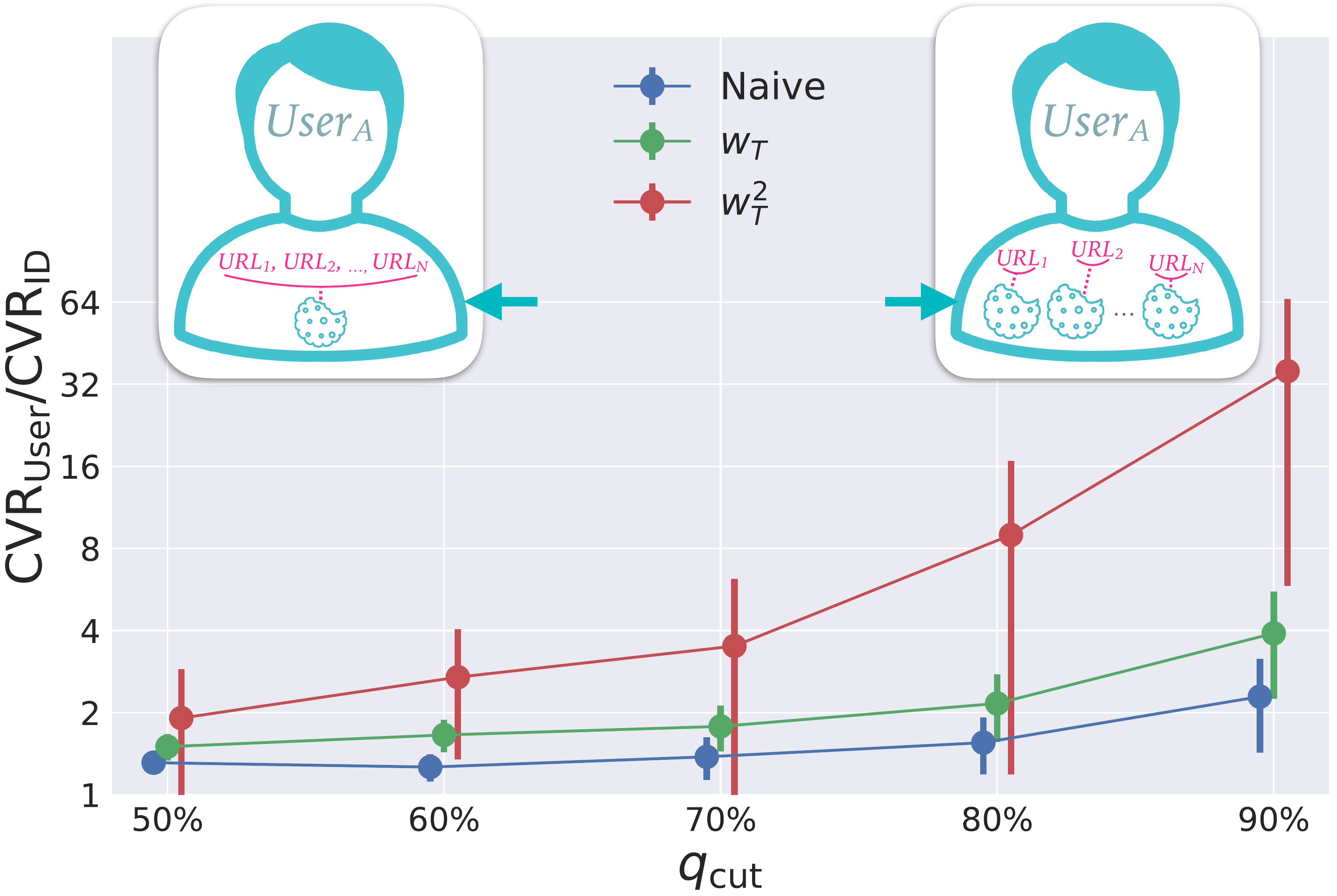}
\caption{CVR ratio of identity-powered to identity-ignorant lookalike targeting for identifiers with $f_i(q_{\textrm{cut}})\!=\!1$ as a function of $q_\mathrm{cut}$ (c.f. Eqn. \ref{qfilter}). Presented are Naive and $w_T$ or $w_T^2$ corrected results. Insets illustrate the limiting cases of low and high $q_{\textrm{cut}}$, where most information is contained on few long-lived identifiers and where information is spread across a variety of short-lived identifiers, resp. Results  are averaged over campaigns and identity dataset $G_1$ is used.}
\label{fig:UserModelResult2}
\end{figure}

An even more interesting finding is presented in Fig.~\ref{fig:UserModelResult2}. Here we show that the advantage of the identity-powered system over the identity-ignorant system is indeed much higher where it is logically expected to be: when the observations on each \emph{identifier} are infrequent, but where more information from the other identifiers associated to the same \emph{user} can supplement. To produce Fig.~\ref{fig:UserModelResult2} we define a boolean filter that is a good proxy for these statements:
\begin{equation}
    f_i(q_\mathrm{cut}) = \big(\mathrm{obs}_i<Q_{\mathrm{obs}}(1-q_\mathrm{cut})\big) \land \big(\mathrm{cls}_i>Q_{\mathrm{cls}}(q_\mathrm{cut})\big),
\label{qfilter}
\end{equation}
where obs$_i$ and cls$_i$ are the number of observations on identifier $i$ and the cluster size of the associated cluster, $Q_{\mathrm{obs}}$ and $Q_{\mathrm{cls}}$ are quantile functions for observations and user cluster sizes\footnote{For reference, $\sim\!\!98\%$ of previously-converted users and $\sim\!\!87\%$ of users considered for Lookalike whitelisting map to non-trivial (multi-vertex) clusters in our base identity dataset $G_1$.}. $f_i(q_\mathrm{cut})$ can thus be used as a boolean filter for including identifier $i$ given a quantile cut $q_\mathrm{cut}$. From Fig. ~\ref{fig:UserModelResult2}, we observe that the CVR lift increases strongly with $q_{\mathrm{cut}}$, reaching values in the $\sim\!(4\!-\!32)$x range for the extreme case $q_{\mathrm{cut}}=90\%$ (amounting to $\sim\!1\%$ of identifiers in our data). The implication is that identity data is especially important in the limit of frequently-churned identifiers.

\section{Discussion}

In this work, we investigated the effectiveness of probabilistically-constructed digital identity data in lookalike-targeted ad campaigns. This data takes the form of hierarchical collections of identifiers, represents identities at a variety of resolutions such as physical devices and users, and are mined from a stream of unstructured and thin event tuples using probabilistic machine learning approaches. We employed off-policy evaluation techniques and a careful handling of the bias that occurs in finite-sized datasets, in order to accurately estimate the lift due to identity data in identity-powered systems above the baseline performance of identity-ignorant variants. Our results suggest a significant lift overall and lend guidance in terms of tuning the identity data being used for this purpose. The lift due to using identity data appears particularly large where it logically should: in the limit where observations associated to an underlying user are spread across a wide number of identifiers that must be inferred to belong to that user (via identity data). 

There are many directions in which this work could be fruitfully extended. Perhaps most obvious is that it would be good to compare these results to those coming from online A/B tests. Unfortunately, this was not possible here because the advertising platform collecting the campaign data became unavailable soon after the data was collected. It would also be interesting to bolster the analysis by including many more campaigns and presenting a fuller analysis of the heterogeneity of the results across campaigns. Here we did not observe a consistent pattern with respect to campaign-specific factors (e.g., call-to-action-type, etc.) among the eight campaigns studied. 

\begin{acks}
  The authors thank Azad Jacobs, Mike Murphy, Shadi Rasheed, Kamakshi Sivaramakrishnan and Jenny Zhao for helpful discussions.
\end{acks}

\bibliographystyle{ACM-Reference-Format}
\balance
\bibliography{sample-bibliography} 

\end{document}